%% file: acl2023.tex
\pdfoutput=1

\documentclass[11pt]{article}

\usepackage[]{ACL2023}
\usepackage{forest}
\usepackage{times}
\usepackage{latexsym}
\usepackage{graphicx}
\usepackage{drs}
\usepackage{tikz}
\usepackage{algorithm}
\usepackage{algorithmic}
\usepackage{dot2texi}
\usepackage[normalem]{ulem} 
\usetikzlibrary{trees, positioning, shapes.geometric}
\usetikzlibrary{shapes.geometric, arrows, positioning}
\usepackage{multirow}
\usepackage{amssymb}
\usepackage{caption}
\usepackage{subcaption}
\usepackage{booktabs}
\usepackage{color, colortbl}
\definecolor{mylightgray}{gray}{0.9} 

\usepackage{tabularray}
\UseTblrLibrary{booktabs}
\usepackage{drs}
\usepackage{latexsym}
\definecolor{Yellow}{rgb}{1.0,1.0,0.0}	
\definecolor{Gray}{gray}{0.9}

\usepackage[first=0,last=9]{lcg}

\usepackage{caption}
\usepackage{multicol}
\usepackage{rotating}
\usepackage{soul}
\usepackage{geometry}
\usepackage{tabularx}
\usepackage{pdflscape}
\usepackage{setspace}
\usepackage{forest}
\usepackage{adjustbox}
\usepackage{cancel}
\usepackage{rotating}
\usepackage{drs}
\usepackage{amssymb}
\usepackage{tcolorbox}
\usepackage{acro}
\usepackage{tikz-dependency}
\usepackage{fancyvrb}
\usepackage{graphicx}
\usepackage{multirow}
\usepackage{subcaption}
\usepackage{color}
\usepackage[T1]{fontenc}

\usepackage[utf8]{inputenc}

\usepackage{microtype}

\usepackage{inconsolata}

\usepackage{todonotes}

\newcommand{\grph}[1]{\textsc{g\textsubscript{#1}}}
\newcommand{\src}[1]{\textsc{#1}}
\newcommand{\MOD}[1]{\textsc{mod\textsubscript{#1}}}

\title{Scope-enhanced Compositional Semantic Parsing for DRT}

\author{
    Xiulin Yang \\
    Georgetown University \\
    \texttt{xy236@georgetown.edu} 
    \And
    Jonas Groschwitz \\
    University of Amsterdam \\
    \texttt{j.d.groschwitz@uva.nl}
    \AND
    Alexander Koller \\
    Saarland University \\
    \texttt{koller@coli.uni-saarland.de} 
    \And
    Johan Bos \\
    University of Groningen \\
    \texttt{johan.bos@rug.nl}
}

\begin{document}

\maketitle


\begin{abstract}

Discourse Representation Theory (DRT) distinguishes itself from other semantic representation frameworks by its ability to model complex semantic and discourse phenomena through structural nesting and variable binding. 
While seq2seq models hold the state of the art on DRT parsing, their accuracy degrades with the complexity of the sentence, and they sometimes struggle to produce well-formed DRT representations.
We introduce the AMS parser, a compositional, neurosymbolic semantic parser for DRT. It rests on a novel mechanism for predicting quantifier scope.
We show that the AMS parser reliably produces well-formed outputs and performs well on DRT parsing, especially on complex sentences.\footnote{The code is accessible via \url{https://github.com/xiulinyang/compositional_drs_parsing}.}

\end{abstract}

\input{introduction}
\input{background}

\input{relatedWork}

\input{method}

\input{results}
\section{Conclusion and Future Work}
\label{conclusion} 

In this work, we proposed a novel mechanism for predicting scope assignments in DRT parsing. By combining it with the compositional AM parser, we obtain the AMS parser, which outperforms existing DRT parsers trained on the same dataset, especially for complex sentences. We show that AMS parser can parse the expressive meaning representation frameworks more easily than the seq2seq counterparts. It can be naturally applied to more expressive meaning representations and our scope resolver can be easily combined with all compositional parsers to parse non-compositional information. Our parser also avoids the prediction of ill-formed DRGs that plague other models. The prediction of scope information has been a long-standing challenge in computational semantics; our dependency parsing mechanism achieves very high accuracy on this task.


In the future, we plan to extend our work to tackle increasingly complex meaning representation frameworks, such as Uniform Meaning Representation (UMR) \citep{van2021designing}. Since UMR-writer \citep{zhao-etal-2021-umr}, the UMR annotation tool, provides node-token alignment automatically, no more manual annotation is needed.  Furthermore, our current system's architecture, which includes both the AM Parser and a dependency parser by \citet{dozat-manning-2018-simpler}, presents opportunities for optimization. We aim to streamline the process by unifying these two models into a single framework that leverages joint learning.

\section*{Limitations}
The AMS parser uses the AM parser to predict the predicate-argument relations in the DRGs. The AM parser has not kept pace in accuracy with the development of overall graph parsing models since it was published in 2019. This holds back the accuracy of the AMS parser. If a more accurate sentence-to-graph parser that induces node-token alignments became available, the AMS parser could be combined with it for increased accuracy. Note, however, that the AM parser shows strong performance with respect to the degradation of parsing accuracy for long and complex sentences.

Furthermore, the treatment of coreference in the paper is quite shallow. One might include the predictions of a coreference resolver into the parsing process. On the relatively short coreference chains in the PMB test sets, this would probably not make a significant impact on the evaluation.

\section*{Acknowledgement}
We would like to express our sincere gratitude to Xiao Zhang, Wessel Poelman, and Chunliu Wang for their generous support and for addressing numerous questions related to DRT parsing throughout this project. This work was made possible through the support of the Erasmus Mundus Master’s program in Language and Communication Technologies.

\bibliography{custom}
\bibliographystyle{acl_natbib}

\input{appendix}

\end{document}

%% file: introduction.tex
\section{Introduction}

\begin{figure}
    \centering
\begin{minipage}{0.2\textwidth}
        \centering
        \scalebox{0.42}{
            \drs{ }{\ifdrs{x}{child.n.01(x)}{e, t, s}{time.n.08(t)\\
            t=now\\
            misbehave.v.01(e)\\
            Manner(e, s)\\
            Time(e, t)\\
            Agent(e, x)\\
            occasionally.r.01(s)}}
        }
    \end{minipage}
\includegraphics[width=0.5\textwidth]{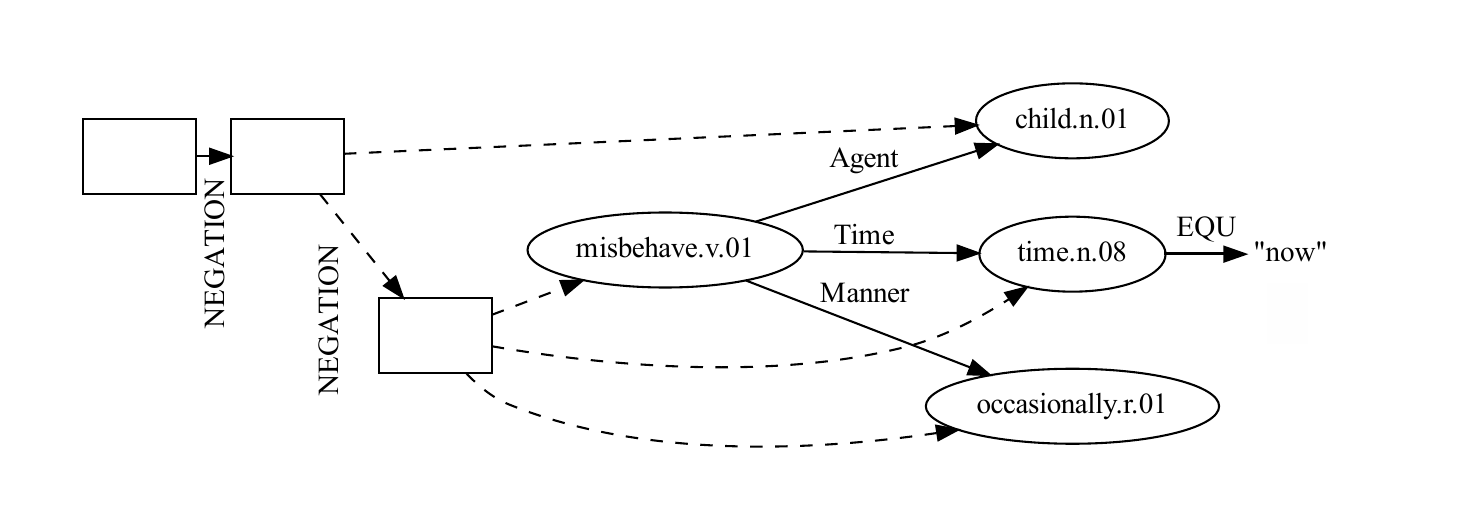}
     \caption{DRS (top) and DRG (bottom) for the sentence \textit{Every child misbehaves occasionally}; dashed lines represent scope assignments of connectives.}
    \label{fig:drsdrg}
\end{figure}

Among current semantic representation formalisms used in NLP, 
Discourse Representation Theory \citep[DRT; ][]{KampReyle1993}
stands out in its systematic use of structural nesting and variable binding to represent meaning in detail. Originating from linguistic theory, DRT has been designed to capture subtle semantic and discourse phenomena such as anaphora, presupposition, and discourse structure, as well as tense and aspect (see  Fig.~\ref{fig:drsdrg}). This structural and semantic richness distinguishes DRT from other popular frameworks in semantic parsing, such as Abstract Meaning Representation \citep[AMR; ][]{banarescu2013abstract}.

With the availability of the broad-coverage Parallel Meaning Bank \citep[PMB; ][]{abzianidze-etal-2017-parallel}, DRT has become an active target for the development of semantic parsing methods. The current state of the art is held by purely neural seq2seq models \citep{zhang-etal-2024-gaining}. However, due to the structural complexity of typical DRT representations, these models do not always generate well-formed meaning representations. They also struggle on long sentences; length generalization is a known challenge for transformers in semantic parsing settings \citep{hupkes2020compositionality,yao-koller-2022-structural}. Existing compositional semantic parsers for DRT significantly lag behind the seq2seq models in terms of parsing accuracy.

In this paper, we introduce the \emph{AMS parser}, an accurate compositional DRT parser. The AMS parser extends the AM parser \citep{groschwitz-etal-2018-amr}, which predicts meaning representations compositionally and has achieved high accuracy across a range of sembanks \citep{lindemann-etal-2019-compositional, weissenhorn-etal-2022-compositional}. The AM parser by itself struggles to predict structural nesting in DRT. The key challenge is to predict \emph{scope}: how to assign each atomic formula in Fig.~\ref{fig:drsdrg} (top) to one of the three boxes which represent scope at different levels. These boxes, which will be explained further in Section~\ref{drt_theory}, play a key role in organizing the logical relationships in DRT. More specifically, in DRT, these boxes, or scope, govern where discourse referents (like individuals or events) and logical operators (such as negation and quantifiers) are valid or accessible within the structure. This is essential for understanding the meaning of sentences, especially in NLP tasks that rely on correct interpretation of negation, quantification, and discourse relations.

The technical contribution of this paper is to extend the AM parser with an innovative mechanism for predicting scope. We train a dependency parser to predict scope relations between word tokens and project this information into the DRT representation using word-to-box alignments. 
We show that this dependency mechanism can predict correct scope assignments at very high accuracy. The overall parser always predicts well-formed DRT representations (in contrast to all seq2seq models) and is almost on par with the best models in parsing accuracy. On the PMB TestLong split, which contains particularly long sentences, it outperforms all other DRT parsers that are trained on the PMB gold dataset. Thus, the strength of the AMS parser is its ability to remain accurate as sentences grow complex.

%% file: background.tex
\section{Background and Related Work}
\label{background}
\subsection{Discourse Representation Theory}
\label{drt_theory}
Discourse Representation Theory \citep[DRT;][]{KampReyle1993} is a well-developed framework for dynamic semantics that aims to interpret meanings from the context. It can model diverse linguistic phenomena ranging from anaphora \citep{kamp1981evenements, haug2014partial} to rhetorical structures \citep{lascarides2007segmented}. 
In DRT, meanings are traditionally represented by Discourse Representation Structures (DRS), which are composed of nested boxes that contain discourse referents (the entities talked about in the discourse) and propositions about these discourse referents.  Fig.~\ref{fig:drsdrg} (top) is an example of DRS representing \textit{Every child misbehaves occasionally}. The boxes act as logical quantifiers that bind variables, and they can be connected with logical operators such as implication.

\citet{bos-2023-sequence} recently proposed an equivalent, variable-free notation for DRSs in the form of directed acyclic graphs, called Discourse Representation Graphs (DRGs; see Fig.~\ref{fig:drsdrg}, bottom). A DRG contains nodes representing boxes, predicate symbols, and constants. Some edges (drawn solid in Fig.~\ref{fig:drsdrg}) connect predicates to arguments with semantic roles. Others (drawn dashed) represent the structural nesting of boxes and propositions:
A dashed edge means that its target node is inside the box from which the edge emanates. Universal quantification, disjunction, and implication are represented in DRGs as logically equivalent structures using only negation and conjunction (see \citet{bos-2023-sequence}).




The main resource for DRS and DRG is the Parallel Meaning Bank~(PMB; \citet{abzianidze-etal-2017-parallel}), which is a multilingual parallel corpus comprising sentences and texts paired with meaning representations. In this paper, we use the latest version (PMB release 5.1.0, English) for evaluation.
It includes three distinct splits based on the quality and method of annotation: Gold (manually verified), Silver (partially corrected), and Bronze (automatically generated by Boxer). As our objective is to address challenges within a limited data setting, our experiments specifically focus on utilizing gold-annotated data.

\subsection{DRS parsing}
\label{review}
Deriving DRSs from sentences compositionally is a nontrivial challenge. Efforts towards this goal  include $\lambda$-DRT \citep{4b5f05af7b974c7c9afd9995ce472e5b,kohlhase1996type, Kohlhase1998DynamicLambda}, Compositional DRT \citep{muskens1996combining}, and bottom-up DRT \citep{asher1993reference}. All of these approaches use lambda calculus to compositionally combine partial meaning representations, which is intractable in broad-coverage semantic parsing (see e.g.\ the discussion by \citet{artzi-etal-2015-broad}).


To date, the most accurate broad-coverage DRT parsers are based on neural sequence-to-sequence models \citep[e.g.,][]{liu-etal-2018-discourse, fancellu-etal-2019-semantic, van2018exploring, van-noord-etal-2020-character}. They achieve impressive performances, especially when the models are trained on additional silver or bronze training data  \citep{wang-etal-2023-pre} or use additional features \citep{van-noord-etal-2019-linguistic, van-noord-etal-2020-character}. However, due to the structure-unaware design of these models, they sometimes struggle to generate well-formed DRT representations (see \citet{poelman-etal-2022-transparent}).

Existing compositional semantic parsers for DRT rely on syntactic dependency parsers
\cite{le-zuidema-2012-learning, poelman-etal-2022-transparent} or CCG parsers \cite{bos-2008-wide, bos2015open}. These models reliably generate well-formed DRSs, but are not competitive with seq2seq models in terms of parsing accuracy.

\subsection{AM Parsing} 

The DRT parser we present here is based on the AM Parser \citep{groschwitz-etal-2018-amr}, a neurosymbolic compositional semantic parser that has previously been shown to be fast and accurate both on broad-coverage parsing, e.g.\ on AMR \citep{lindemann-etal-2019-compositional}, and in compositional generalization tasks \citep{weissenhorn-etal-2022-compositional}.

\paragraph{Apply and Modify}
The AM parser uses a neural dependency parser and tagger to predict terms over the AM algebra \citep{groschwitz-etal-2017-constrained}, which combines graphs into bigger graphs using the operations \textit{Apply} and \textit{Modify}. To this end, nodes of the graphs can be decorated with \emph{sources} (\citet{courcelle2012graph}, marked in blue), which assign names to nodes at which the graph can be combined with other graphs. Every graph has a special source called \src{root}, drawn with a bold outline, which is where the graph inserts into others when used as an argument.

\begin{figure}
    \centering
     
    \begin{subfigure}[b]{0.35\textwidth}
        \centering
        \includegraphics[width=\textwidth]{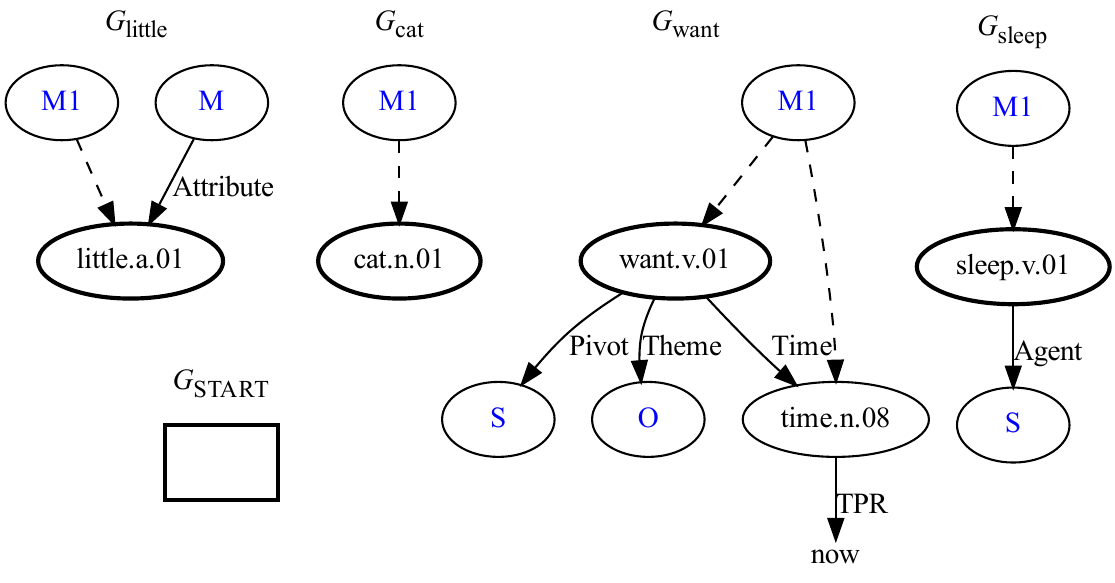}
        \captionsetup{font=scriptsize}
        \caption{Graphs for the individual words}
        \label{fig:0}
    \end{subfigure}
    \hfill
     \begin{subfigure}[b]{0.11\textwidth}
        \centering
\includegraphics[width=0.85\textwidth]{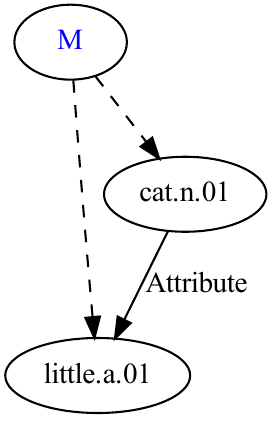}
        \captionsetup{font=scriptsize}
\caption{\textsc{mod\textsubscript{m}}\textsc{(g}\textsubscript{cat}, \textsc{g}\textsubscript{little})}

        \label{fig:3}
    \end{subfigure}
    \begin{subfigure}[b]{0.27\textwidth}
        \centering
\includegraphics[width=0.87\textwidth]{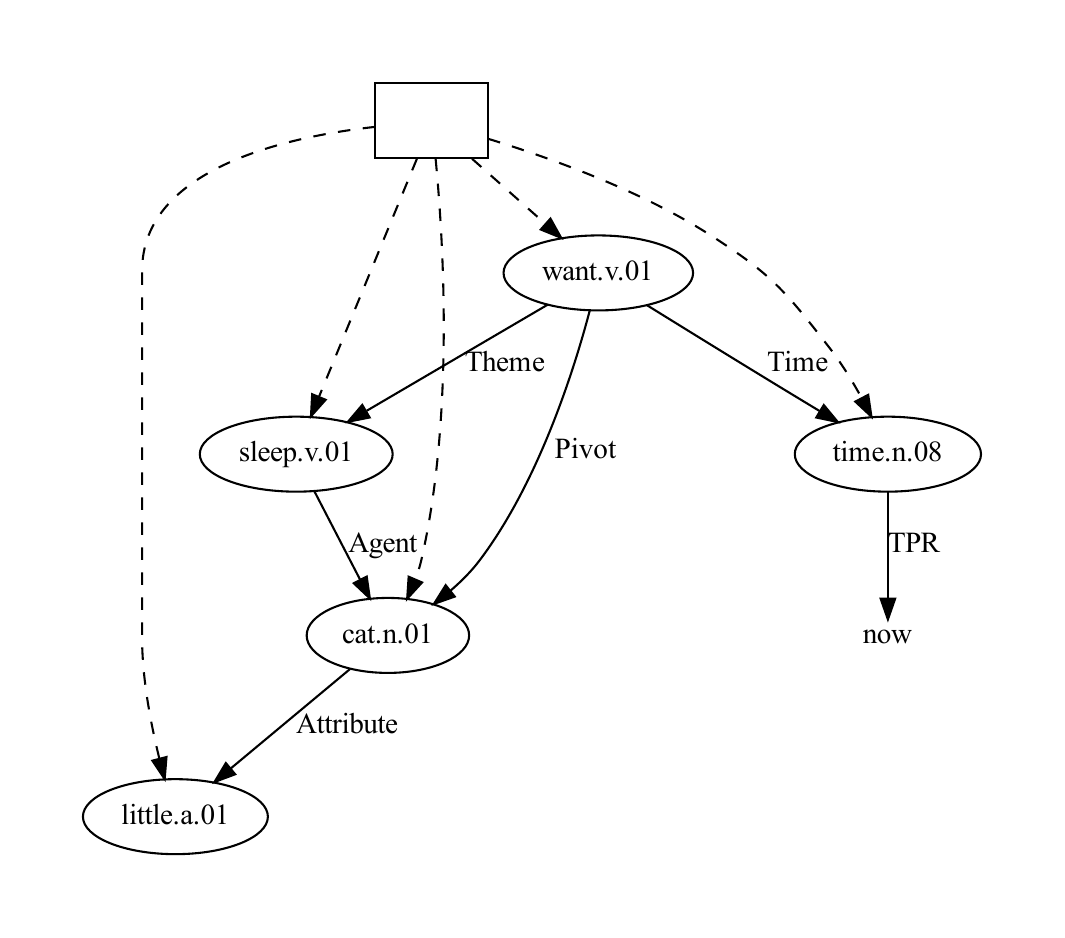}
       \captionsetup{font=scriptsize}
        \caption{Complete DRG}
        \label{fig:1}
    \end{subfigure}
    \hfill
    \begin{subfigure}[b]{0.18\textwidth}
        \centering
    \includegraphics[width=0.9\textwidth]{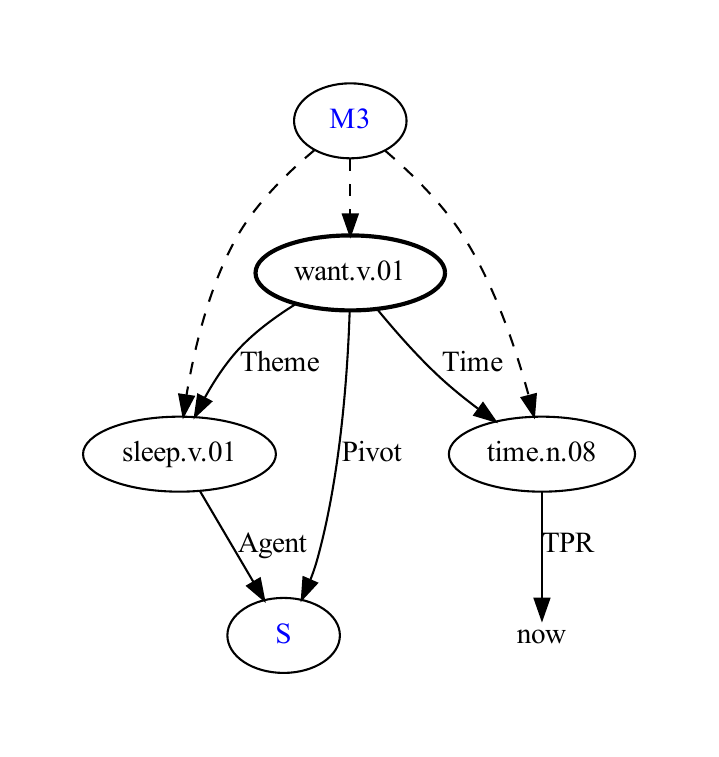}
        \captionsetup{font=scriptsize}
    \caption{\textsc{app}\textsubscript{\textsc{o}}(\textsc{g}\textsubscript{want}, \textsc{g}\textsubscript{sleep})}

        \label{fig:2}
    \end{subfigure}
        
     \begin{subfigure}[b]{0.38\textwidth}
    \begin{dependency}[theme = simple]
   \begin{deptext}[column sep=.01cm, font=\tiny]
    \textsc{start} \& The \& little \& cat \& wanted \& to \& sleep. \\
    \colorbox{green!20}{\textsc{g}\textsubscript{\textsc{start}}} \& \& \colorbox{red!20}{\textsc{g}\textsubscript{little}} \&   \colorbox{yellow!20}{\textsc{g}\textsubscript{cat}} \&  \colorbox{blue!20}{\textsc{g}\textsubscript{wanted}} \&  \& \colorbox{orange!20}{\textsc{g}\textsubscript{sleep}} \\
   \end{deptext}
   \depedge{1}{5}{\colorbox{green!20}{\textsc{MOD}\textsubscript{m}}}
   \depedge{5}{4}{\colorbox{blue!20}{\textsc{app}\textsubscript{s}}}
   \depedge{5}{7}{\colorbox{blue!20}{\textsc{app}\textsubscript{o}}}
   \depedge{4}{3}{\colorbox{yellow!20}{\textsc{mod}\textsubscript{m2}}}
\end{dependency}
\captionsetup{font=scriptsize}
\caption{AM dependency tree}
\label{fig:amdeptree}
    \end{subfigure}
    




    
    \caption{Relevant graphs for sentence \textit{The little cat wanted to sleep.}}
    \label{fig:composite}
\end{figure}

In the example of Fig.~\ref{fig:0}, the graph \grph{want} has sources \src{s} and \src{o} indicating where the arguments supplied by the subject and object should be inserted. It also has a source \src{m1} which allows it to attach to some other graph as a modifier.




The \textit{Apply} operation (\textsc{app}) models the combination of a complement (i.e.~argument) with its head. For example in Fig.~\ref{fig:2}, the \textsc{app\textsubscript{o}} operation combines the head \textsc{g\textsubscript{want}} with its argument \textsc{g\textsubscript{sleep}}, plugging the root of \textsc{g\textsubscript{sleep}} into the \textsc{o} source of \textsc{g\textsubscript{want}} (Fig.~\ref{fig:2}). Because every graph may only contain one node decorated with each source name, the \textsc{s} and \textsc{m1} source nodes of \textsc{g\textsubscript{sleep}} and \textsc{g\textsubscript{want}} get merged. This allows the AM algebra to generate nontrivial graph structures.



The \textit{Modify} operation (\textsc{mod}) models the combination of a head with a modifier. For example, the \textsc{mod\textsubscript{m}} operation in our example attaches the adjunct \textsc{g\textsubscript{little}} to the root of its head \textsc{g\textsubscript{cat}}, using the adjunct's \textsc{m} source (Fig.~\ref{fig:3}). Again, both graphs have an \textsc{m1} source that gets merged.

\paragraph{AM dependency trees and AM parsing}
The AM parser predicts a graph from a sentence by computing an \emph{AM dependency tree}, as in Fig.~\ref{fig:amdeptree}. It uses a neural tagger to predict a \emph{lexical graph} for each word (drawn below the sentence) and a neural dependency parser to predict \textsc{app} and \textsc{mod} edges. The AM dependency tree can be unraveled into a term of \textsc{app} and \textsc{mod} operations over the AM algebra, which deterministically evaluates into a graph; for instance, the AM dependency tree in Fig.~\ref{fig:amdeptree} evaluates to the graph in Fig.~\ref{fig:1}. Words that do not lexically contribute to the meaning representation, such as the determiner \textit{the}, are not assigned incoming dependency edges and thus ignored in the construction of the graph.




In order to train the AM parser, one needs to construct an AM dependency tree for every sentence-graph instance in the training data. \emph{Decomposing} the graph into an AM dependency tree is a nontrivial task, which can fail: Depending on the alignments between word tokens and nodes in the graph, an AM dependency tree that evaluates to the given graph may not exist. We call such training instances \emph{non-decomposable}.



%% file: relatedWork.tex
\section{Scope in DRT is hard for the AM parser}
\label{sec:challenge}

We start with an attempt to directly apply the AM parser to DRT.
As we will see, the dashed scope edges in a DRG are difficult to handle with the AM parser. We will solve this problem in the AMS parser, presented in Section~\ref{method}.


\subsection{A baseline AM parser for DRG}
We construct AM dependency trees for the DRGs in the PMB using the
approach proposed by 
\citet{groschwitz-etal-2021-learning}, which learns decomposition jointly with training the neural parsing model. The learning algorithm represents the latent space of possible AM dependency trees for each graph compactly, allowing training on the whole latent space. This leads to the parser converging on AM dependency trees that are consistent across the corpus.

This largely unsupervised method still requires two inputs beyond the graph. First, node-token alignments (every node must be aligned), for which we use the alignments given in PMB5.1. For the top box in each DRG, which is always unaligned, we introduce a special \texttt{START} token to align it to (cf.~Fig.~\ref{fig:amdeptree}).

Second, each edge must be assigned to a graph constant, to fully partition the DRG into lexical graphs for the individual words. This often makes the difference between an Apply or Modify operation. For example in Fig.~\ref{fig:composite}, the \texttt{Attribute} edge between \texttt{little} and \texttt{cat} is grouped with the \texttt{little} node, making \texttt{little} a modifier of \texttt{cat}, a linguistically plausible analysis. The edge could also be grouped with the \texttt{cat} node, effectively making \texttt{little} an argument of \texttt{cat} (the two would be combined with an \textsc{app} operation), an implausible analysis. We follow the linguistically-informed principle to group edges between a head and an argument with the head, and edges between a head and a modifier \citep{lindemann-etal-2019-compositional}; see Appendix~\ref{direction} for our full heuristics. The scope edges do not fall into these categories and provide a unique challenge, see below.

All remaining aspects of the AM dependency tree, including the source names, are then learned during training.


 
\subsection{The Challenge of Scope Prediction}
\label{challenge}

\begin{figure}
    \centering
\includegraphics[scale=0.4]{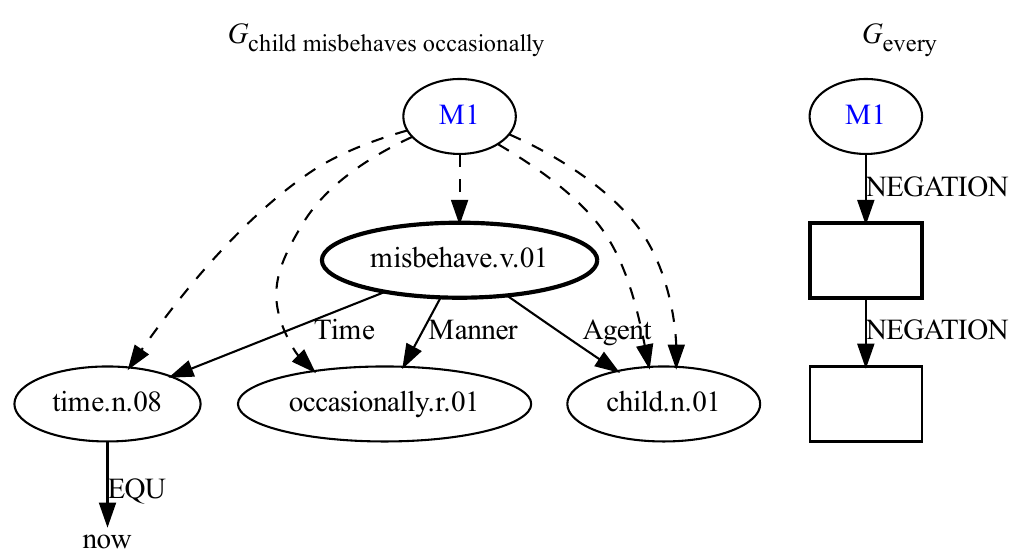}


    \caption{Failed combination of graphs for Fig~\ref{fig:drsdrg}}
    \label{fig:failcomposite}
\end{figure}
The scope edges of DRGs are not something that the Apply and Modify operations were designed for. In particular, the scope edges do not fall straightforwardly into the head/argument/modifier paradigm. The design of the AM algebra forces us into an inconvenient choice: (1) include scope edges in the lexical graph that contains the box and insert the contents of the box with Apply operations; or (2) include scope edges in the lexical graphs of the contents of the box and insert them into the box using Modify operations.

\begin{table}[t]
    \small
    \centering
    \begin{tabular}{lcccc}
    \toprule
         & NoPrep & CMPT & SCPL \\
    \midrule
    APP & 0.7 & 84.5 & 94.4  \\
    MOD & 76.7 & 77.7 & 78.0 \\
    \bottomrule
    \end{tabular}
    \caption{Decomposable graphs in PMB5 (\%). APP: member edges grouped with the box; resulting in Apply operations in the AM dependency tree. MOD: member edges grouped with the content nodes, resulting in Modify operations.
    }
    \label{tab:decomposable_stat}
    \end{table}

The first approach fails completely, with only 0.7\% of DRGs in PMB5 being decomposable (see Table~\ref{tab:decomposable_stat}, NoPrep/APP; see also Appendix~\ref{scope_challenge}). The second approach works better, with 76\% graphs being decomposable (Table~\ref{tab:decomposable_stat}, NoPrep/MOD). For example, Fig.~\ref{fig:amdeptree} shows a valid AM dependency tree for the graph in Fig.~\ref{fig:1} under this paradigm. However, this success is limited to graphs with only a single box: only 30\% of all multibox DRGs, i.e.\ DRGs that contain more than one box node, can be decomposed into AM dependency trees. 

To illustrate the challenge, consider the DRG in Fig.~\ref{fig:drsdrg}. Fig.~\ref{fig:failcomposite} shows two partial graphs in an attempt to build the full graph with the AM algebra, the left representing \textit{child misbehaves occasionally}, and the right representing \textit{every}. The lexical graph \grph{every} introduces two boxes, and to obtain the DRG in Fig.~\ref{fig:drsdrg}, we need to draw a scope edge from the upper box to the \textit{child} node on the left and, simultaneously, scope edges from the lower box to the \textit{misbehave}, \textit{time}, and \textit{occasionally} nodes. We can use a \MOD{m1} operation to unify the \src{M1}-source of the left graph with the root of \grph{every} (the upper box); but this will put \textit{child} into the wrong box. The problem is that both boxes are introduced by the same lexical graph (a consequence of the alignments in the PMB), and only one of them can receive outgoing edges through a single Modify operation. Other attempts at decomposing the DRG in Fig.~\ref{fig:drsdrg} fail in similar ways.

%% file: method.tex
\section{Scope-enhanced AM Parsing}
\label{method}

\begin{figure}[t]

    \centering
    \includegraphics[scale=0.6]{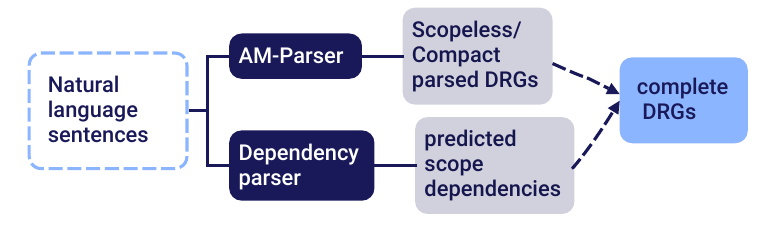}
    \caption{Overall structure of the AMS parser.}
    \label{fig:pipeline}
\end{figure}

We will address this scope challenge through a two-step process. First, we simplify the DRGs by removing scope edges, such that over 94\% of DRGs can be decomposed for training. 
Second, we recover the scope information at parsing time through an independent scope prediction mechanism.  The overall structure of our parser is sketched in Fig.~\ref{fig:pipeline}.

%

\subsection{Simplifying DRGs}\label{subsec:simplifyingDRGs}
We identified two effective DRG simplification strategies: \textit{Compact DRG} and \textit{Scopeless DRG}. 

\begin{figure}[t]
    \centering
    \includegraphics[scale=0.36]{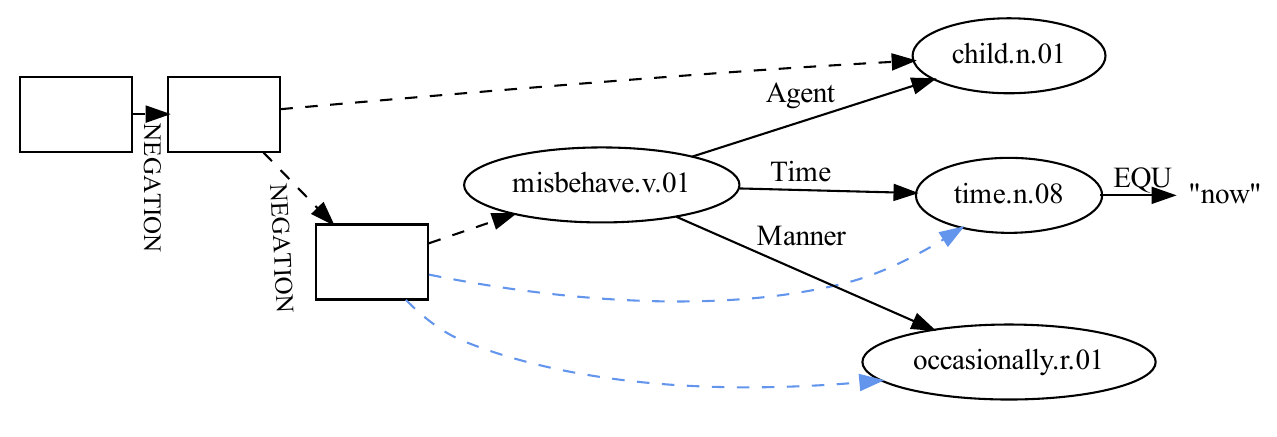}
    \caption{Compact DRG for Fig~\ref{fig:drsdrg} (The removed edges are marked in light blue). }
    \label{fig:compactdrg}
\end{figure}

\paragraph{Compact DRG} The Compact DRG representation (CPT), inspired by \citet{abzianidze-etal-2020-drs}, makes use of the fact that many nodes share the same scope as their parent node, i.e.~are members of the same box. In this representation, we thus remove all scope edges for nodes that are in the same scope as their parents (if there are multiple parents, we only remove the scope edge if the node and \emph{all} its parents are in the same box). This method removes around 70\% of scope edges, and the full scope information can be losslessly recovered with the rule-based method in Section~\ref{subsec:scopeResolution}. The compact DRG for Fig.~\ref{fig:drsdrg} is shown in Fig.~\ref{fig:compactdrg} with the removed edges marked in light blue.
 \paragraph{Scopeless DRG} While Compact DRGs maintain at least one connection between a scope box and a node within its scope, Scopeless DRGs (SCPL) remove all scope edges as long as the graph remains connected. This results in graphs that are mostly reduced to their predicate-argument structure, facilitating a more straightforward decomposition with the AM Algebra, at the cost of losing some information. An example is shown in Fig.~\ref{fig:scopeless-exp}. More complex examples are detailed in Appendix~\ref{simples}.

 \begin{figure}[t]
     \centering
     \includegraphics[scale=0.35]{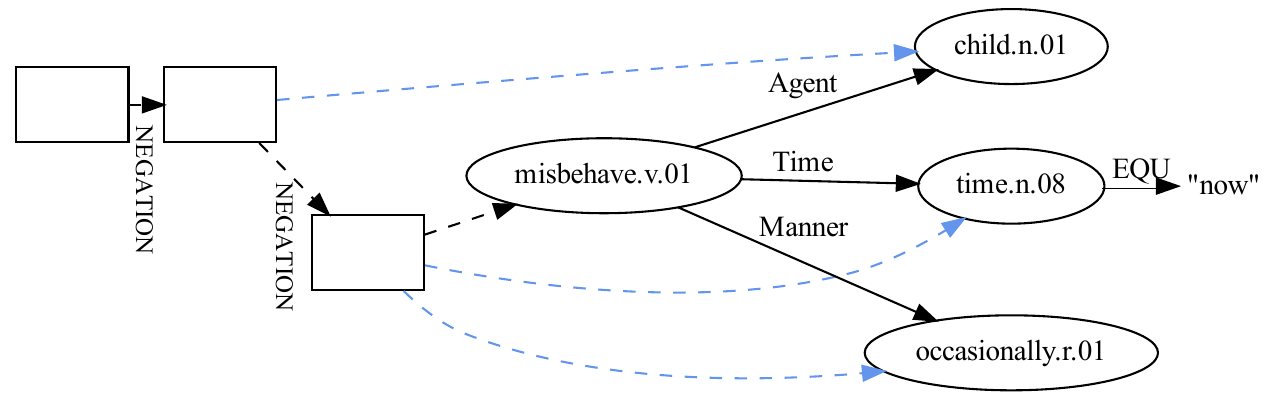}
     \caption{Scopeless DRG for Fig~\ref{fig:drsdrg} (The removed edges are marked in light blue).}
     \label{fig:scopeless-exp}
 \end{figure}
 Both Compact and Scopeless DRGs show much higher decomposability rates compared to the full DRGs, see Table~\ref{tab:decomposable_stat}. This effect is particularly strong in the setting where membership edges are grouped with the boxes (see row ``APP''), where Compact and Scopeless DRGs achieve decomposability rates of 84.5\% and 94.4\% respectively.
 


\subsection{Scope Prediction}\label{subsec:scopeResolution}
To recover the scope information, we designed two scope resolvers: one rule-based, and the other relying on a dependency parser to predict the scope edges.

\paragraph{Rule-based Scope Resolver}
The rule-based scope resolver is the inverse of our Compact DRG simplification method, but can also be applied to Scopeless DRG. This resolver 
traverses the predicted graph top-down; if it encounters a node with no incoming scope edge, it assigns the node the same scope as its parent.
If a node has multiple parents with conflicting scope, an arbitrary parent is chosen (this only occurs with Scopeless DRG). For Compact DRG, this method recovers the full scope information losslessly. 


This rule-based approach is easy to implement, transparent and fully explainable. However, it is imperfect for Scopeless DRG, and even for Compact DRG it may propagate parsing errors into the recovered scope edges.





\paragraph{Dependency-based Scope Resolver}
\label{sec:dependency-scope}

\begin{figure}[t]
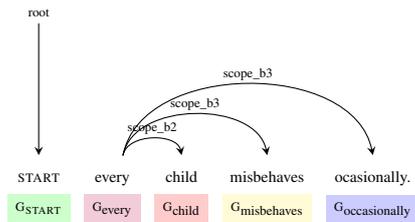

\centering
    \centering
        \begin{dependency}[theme = simple]
   \begin{deptext}[column sep=.01cm, font=\tiny]
   \textsc{start} \& every \& child \& misbehaves \& ocasionally.\\
    \colorbox{green!20}{\textsc{g}\textsubscript{\textsc{start}}} \&  \colorbox{purple!20}{\textsc{g}\textsubscript{every}}\& \colorbox{red!20}{\textsc{g}\textsubscript{child}} \&   \colorbox{yellow!20}{\textsc{g}\textsubscript{misbehaves}} \&  \colorbox{blue!20}{\textsc{g}\textsubscript{occasionally}}\\
   \end{deptext}
   \deproot{1}{root}
   \depedge{2}{3}{scope\_b2}
   \depedge{2}{4}{scope\_b3}
   \depedge{2}{5}{scope\_b3}
\end{dependency}
    \caption{Scope dependency graph for \textit{Every child misbehaves occasionally.} \label{fig:dependency-tree}} 
    \label{fig:example-complex-drg}
\end{figure}

For the dependency-based scope resolver, we make use of the fact that
an AM dependency tree splits the graph into lexical graphs, each of which is linked to a specific word token in the sentence.
This induces an alignment relation between nodes in the graph and tokens in the sentence: a node is aligned to the token if it is part of the lexical graph for that token.
We project the scope edges in the DRG into edges between the word tokens by following this alignment relation from the nodes to the tokens; this creates a  \emph{scope dependency graph} over the sentence (see Fig.~\ref{fig:dependency-tree}). The scope dependency graph is not necessarily a tree: it need not be connected, and a token might receive multiple incoming edges if the aligned lexical graph contains multiple nodes linked to different boxes  (see Appendix~\ref{scope_annotation}).




When the lexical graph for a token contains multiple nodes or boxes, we also encounter a further challenge. In such a case, the scope dependency graph, which connects only the two tokens, cannot fully specify which nodes in the lexical graphs the scope edge connects. An example of this can be seen in Fig.~\ref{fig:dependency-tree}. Here, the lexical graphs \grph{child} and \grph{misbehaves} are both children of \grph{every} in the scope dependency graph, but they should go in different boxes of \grph{every}.

To remove this ambiguity, we name the boxes in each lexical graph and encode the box to which each child in the scope dependency graph connects in the dependency edge label. For example, consider again the dependency graph in Fig.~\ref{fig:dependency-tree}. The relationship between the tokens \textit{every} and \textit{child} is annotated as \texttt{scope\_b2}, indicating that \textit{child} goes into the upper box (\texttt{b2}\footnote{The labeling of the boxes is decided by the hierarchy of the boxes in the whole graph: the parent box is assigned by a smaller number than the children, the root box is assigned with \texttt{b1}.}). By contrast, the edge into \textit{misbehaves} has the label \texttt{scope\_b3}, indicating that it goes into the lower box. We use a similar method if different nodes from the same constant are members of different boxes, see Appendix~\ref{scope_annotation}.
In this way, the labeled scope dependency graphs unambiguously specify scope edges. 

This method allows us to use standard dependency parsing techniques for scope prediction. We adopted the biaffine dependency graph parser of \citet{dozat-manning-2018-simpler}, which is simple and accurate. We use ordinary supervised training, based on the annotated node-token alignments in the PMB5.
Hyperparameter details can be found in Appendix~\ref{appendix:scopeDependencyParser}.


Since the AM parser also predicts some scope edges (in particular for Compact DRG, but also a bare minimum in Scopeless DRG), there can be conflicts between the dependency-based scope predictions and scope edges already present in the predicted simplified DRG. We use the following rules to resolve mismatches: (1) We only use a dependency-based edge if its target has no scope edge in the predicted simplified DRG; i.e. the AM parser predictions take precedence. (2) Any remaining node without a scope edge inherits its scope from its parent (as in the rule-based resolver).

%% file: results.tex
\section{Results \& Discussion}
\label{result}
\subsection{Data \& Evaluation}
We evaluated on the latest Parallel Meaning Bank 5.1.0  \cite{abzianidze-etal-2017-parallel}. Apart from the normal train, dev, and test split, the PMB 5.1.0 also provides an extra TestLong set that contains 40 lengthy (average length: 39.7 tokens) sentences. Statistics can be found in Appendix~\ref{data}. 




For the evaluation metric, we follow \citet{wang-etal-2023-discourse} and convert DRGs to condensed Penman notations\footnote{Examples can be found in Appendix~\ref{evaluation_format}.}. We adopt the SMATCH F1 score \citep{cai-knight-2013-smatch} and SMATCH++ F1 \citep{opitz-2023-smatch} to assess DRGs due to recent findings indicating that the current hill-climber graph-alignment solvers in SMATCH do not ensure fair comparisons. As these metrics are originally designed for AMR and assume an implicit \texttt{root}, which is not applicable in DRG, we customize them by ignoring the root node. We also report the percentage of test instances for which a parser generated ill-formed DRGs. By \textit{ill-formed}, we mean the output sequence cannot be successfully converted to a Penman graph; for instance, an ill-formed graph might assign a node to a non-existent scope.

\subsection{Handling coreference}

The PMB contains coreference annotations; these are non-compositional by design and thus very tricky for a compositional system like the AM parser.
We reduce the impact of coreference on our evaluation through a simple pre- and postprocessing method. We remove all edges indicating coreference in the DRG and introduce a new tag \texttt{p} to the label of all coreferent nodes. In postprocessing, we then simply add coreference edges between all nodes marked as coreferent. This method further increases decomposability, up to 94\% (see Table~\ref{tab:decomposable_stat}).
Details are in Appendix~\ref{coref}.

This method has the advantage of only using information from the predicted DRG, but it only really works when there is just one instance of coreference in the graph. This is frequently the case in the PMB 5.1.0, but in a different setting, more complex coreference resolution methods would likely be needed (see~e.g.~\citet{anikina-etal-2020-predicting}).

\subsection{Experiment details}
We use the implementation of \citet{groschwitz-etal-2021-learning} in all our AM Parser experiments. Hyperparameter settings can be found in Appendix~\ref{amppp}.



We compare the AMS parser against the strongest published models for DRG parsing listed in \citet{zhang-etal-2024-gaining}: \texttt{byT5} \citep{xue-etal-2022-byt5}, \texttt{mT5} \citep{xue-etal-2021-mt5} and \texttt{mBART} \citep{liu-etal-2020-multilingual-denoising}. All of these are sequence-to-sequence models with no built-in awareness of semantic structure, compositionality, or scope.

We also trained the AM Parser on the DRGs with the original scope annotations. To make the root box easier to learn for the parser, we introduced a new token \texttt{START} to the beginning of each input sentence. Finally, we fine-tuned  \texttt{T5-Base}, \texttt{T5-Large} \citep{raffel2020exploring} as two further robust baselines.
 


\subsection{Parsing Results}
\paragraph{Scope Dependency Parsing} 
We first evaluate how accurately scope assignments can be predicted by dependency parsing (cf.\ Section~\ref{sec:dependency-scope}), using the usual UAS and LAS evaluation measures for dependency parsing. Table~\ref{table:edgeparse} reveals high LAS and UAS of predicted scope dependency graphs across both development and test sets, indicating reliable scope prediction. This is remarkable, given the complexity of the scope prediction task. 

On the TestLong set, the accuracy dropped significantly, indicating 
the difficulty of predicting scope as sentences grow in complexity. The much larger drop in LAS compared to UAS indicates the difficulty of reliably making scope assignment decisions within a lexical graph.

\begin{table}[t]
\centering
\footnotesize 
\begin{tabular}{@{}lcccccccccccccccc@{}}
\toprule
 \multicolumn{2}{c}{Dev} & \multicolumn{2}{c}{Test} & \multicolumn{2}{c}{TestLong}\\
\cmidrule(lr){1-2} \cmidrule(lr){3-4}\cmidrule(lr){5-6}
 \multicolumn{1}{c}{UAS} & \multicolumn{1}{c}{LAS} & \multicolumn{1}{c}{UAS} &\multicolumn{1}{c}{LAS} & \multicolumn{1}{c}{UAS} &\multicolumn{1}{c}{LAS}\\
\midrule

 98.7 & 96.4 & 98.3 & 95.7 & 67.0 & 55.4\\
\bottomrule
\end{tabular}
\caption{Accuracy of scope dependency parsing.}
\label{table:edgeparse}
\end{table}

\begin{table*}[t]
  \centering
  \small
  \begin{tabular}{lrrrrrrrrrrr@{}}
  \toprule
  Models & \multicolumn{3}{c}{Test} & \multicolumn{3}{c}{TestLong} \\
  \cmidrule(lr){2-4} \cmidrule(lr){5-7}
   &  F1\textsubscript{SMATCH}& F1\textsubscript{SMATCH++} & Err &F1\textsubscript{SMATCH}  & F1\textsubscript{SMATCH++}  & Err \\
  \midrule
  \midrule
\textit{\textbf{Baselines (gold only)}}\\
  ByT5\textsubscript{(G)}  & 86.7 & \textbf{90.0} & 5.4 & 27.1 & 29.0 & 38.3 \\
   mT5\textsubscript{(G)}  & 61.2 & 67.1 &  11.3 & 16.5 & 17.2 & 25.0 \\
   mBART\textsubscript{(G)}  & 82.8 & 84.3 & 6.3 & 30.5 & 32.7 & 12.5 \\
   T5-base\textsubscript{(G)} & 76.4 & 80.1 & 20.0 & 13.9 & 13.6 & 77.5 \\
   T5-large\textsubscript{(G)} & 84.2 & 88.2 & 3.9 &18.1 & 17.1 & 67.5\\
  AM Parser\textsubscript{(G)} & 83.7 & 85.2 & \textbf{0.0} & 46.6 & 48.4 & \textbf{0.0} \\
  \midrule
  \footnotesize{\textbf{\textit{without scope resolution}}}\\
  AMS Parser\textsubscript{scpl(G)} & 70.1 & 76.8 & \textbf{0.0} & 38.4 & 40.9& \textbf{0.0} \\
  AMS Parser\textsubscript{cmpt(G)} & 71.4 & 76.3 & \textbf{0.0} & 36.5 & 37.9 & \textbf{0.0} \\
  \midrule
  \footnotesize{\textbf{\textit{with scope resolution}}}\\
  AMS Parser\textsubscript{scpl+h(G)}& 86.1 & 88.0 & \textbf{0.0} & 48.7 & 50.0 & \textbf{0.0} \\
  AMS Parser\textsubscript{cmpt+h(G)} & 85.6 & 86.7 & \textbf{0.0} & 44.2 & 44.4 &  \textbf{0.0} \\
  AMS Parser\textsubscript{scpl+d(G)} & \textbf{87.1} & 88.9 &  \textbf{0.0} & \textbf{48.7} & \textbf{50.8} & \textbf{0.0} \\
  AMS Parser\textsubscript{cmpt+d(G)} & 85.8 & 87.0 &  \textbf{0.0} & 45.5 & 47.3 &  \textbf{0.0} \\
  \midrule
  \midrule
  \footnotesize{\textit{\textbf{Baselines (gold + silver)}}}\\
  byT5\textsubscript{(G+S)} & 93.4 &93.9 &  0.7 & 36.6 & 47.3 & 40.0 \\
  T5-base\textsubscript{(G+S)} & 86.0& 90.2 &  1.6 & 44.3 & 44.4 &37.5 \\
  mT5\textsubscript{(G+S)} & 93.1 & 93.7 & 0.8 & 55.8 & 64.4 &  15.0 \\
   mBART\textsubscript{(G+S)} & 86.2 & 88.5 & 4.4 & 7.8 & 13.8 & 12.5 \\
   \bottomrule
  \end{tabular}
  \caption{Accuracy and error rates for DRG parsing.} 
  \label{table:pmb5}
  \end{table*}
\paragraph{DRG Parsing} For the task of DRG parsing itself, we compare the AMS parser to the baselines in Table~\ref{table:pmb5}. Our focus is on models that are trained on the hand-annotated gold dataset (\textit{G}); we also include some models trained on gold and silver. The suffix \textit{scpl} denotes Scopeless DRGs, \textit{cmpt} refers to Compact DRGs, \textit{d} indicates the dependency-based scope resolver, and \textit{h} signifies the heuristic scope resolver. ``Without scope resolution'' groups together variants of the AMS parser that directly predict compact or scopeless DRGs, without a mechanism for reconstructing scope edges in postprocessing. The best results among the gold-trained models are marked in bold. Note that SMATCH++ scores are usually higher than SMATCH, but the overall ranking does not change. Three critical observations emerge from the table. 

First, the AMS parser, especially the scopeless (SCPL) version, excels against the gold-data trained baselines. The only exception is \texttt{byT5}, which has a token-free architecture that makes it particularly good at processing short texts -- a significant advantage given the very short average sentence length of 6.7 tokens in the regular test set. The AMS parser also outperforms the generic AM parser, indicating the effectiveness of our novel scope resolution mechanism.

Second, in contrast to all seq2seq models, the AMS parser maintains a 0\% error rate, i.e.\ it never generated ill-formed DRGs. Furthermore, on the very long sentences of the TestLong set, all variants of the AMS parser outperform the gold-trained seq2seq baselines by a large margin, better than all models trained on silver data except for \texttt{mT5}.


Finally, Scopeless DRGs perform better than Compact DRGs. This could be attributed to the fact that Compact DRGs retain more scope edges, making the graph more complex to learn. The higher decomposability rate of Scopeless DRGs also means that we have more training data in that setting. The dependency-based scope resolver outperforms its heuristic-based counterpart in accuracy across in-domain development and test splits. This advantage makes sense given the scope dependency parser's high accuracy. It could also be that the dependency resolver is better able to handle initial parsing inaccuracies compared to the rule-based resolver, where AM Parsing errors can easily propagate into more scope errors.

\paragraph{Scaling to complex DRGs}

As we already saw in Section~\ref{challenge}, scope prediction is easy when there are not many boxes. Table~\ref{table:multibox_scope} therefore splits the test instances by number of boxes\footnote{For the TestLong split, we evaluate the models only on multi-box DRGs, of which there are 33 out of 40.}. For each of these classes, we report the overall SMATCH score of our best model and the baselines, as well as the SMATCH score when considering only scope edges. In the table, the values highlighted in gray represent the \textit{scope} score (i.e., the SMATCH score specifically for scope predictions), while the values in the white rows show the overall score (i.e., the SMATCH score for the entire graph prediction). This allows us to explore how the parsers scale to complex DRGs, and in particular how they maintain their ability to predict scope edges when there are many boxes. Note that \textit{Count} refers to the number of instances that have the specific number of boxes (\textit{\# Box}). As we can see, as the graphs become more complex, the dataset contains fewer examples.

\begin{figure}
    \centering
    \includegraphics[width=\linewidth]{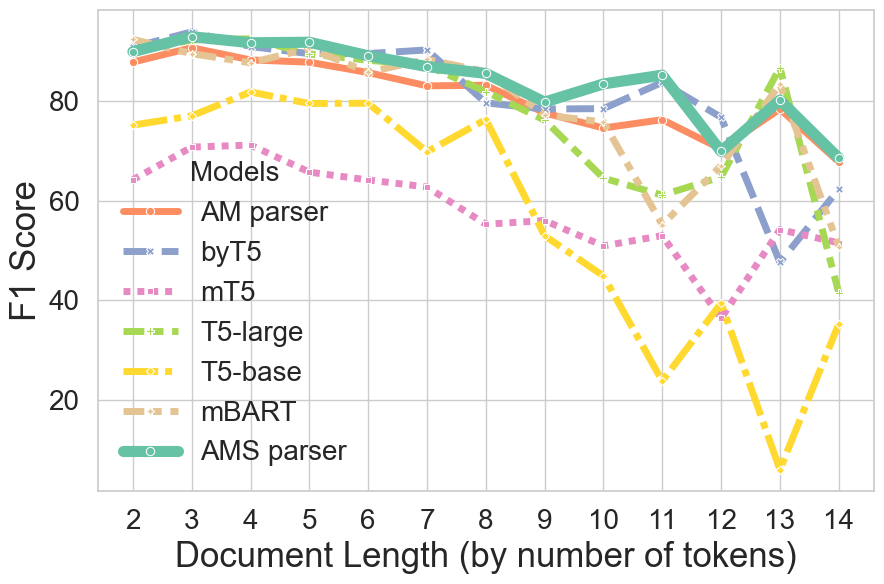}
    \caption{SMATCH++ F1 for DRGs Across Different Models and Document Lengths.}
    \label{fig:drg_length}
\end{figure}

Compared to other models trained on gold data, the AMS parser excels at maintaining its accuracy as the DRGs grow more complex. While the AM parser is almost on par with the AMS parser on single-box DRGs, the gap widens drastically with increasing complexity. For DRGs with four or more boxes, as well as on the TestLong set, the AMS parser also decisively outperforms all (gold) seq2seq baselines. Additionally, as shown in Fig.~\ref{fig:drg_length}, the AMS parser demonstrates more stable performance than the other gold-based models as the sequence length increases.


\begin{table}[t]
\centering
\footnotesize
\begin{tblr}{
  colspec = {lcccccccc},
row{5} = {bg=mylightgray}, 
row{7} = {bg=mylightgray},
row{9} = {bg=mylightgray},
row{11} = {bg=mylightgray}, 
row{13} = {bg=mylightgray},
row{15} = {bg=mylightgray},
row{17} = {bg=mylightgray},
row{19} = {bg=mylightgray},
row{21} = {bg=mylightgray},
row{23} = {bg=mylightgray},
  column{1} = {leftsep=0pt}, 
  column{Z} = {rightsep=0pt}, 
  cells = {halign=c, valign=m}, 
}
\toprule
Models & \SetCell[c=4]{c} Test & & & & \SetCell[c=1]{c} TestL \\
\# Box & \#1 & \#2 & \#3 & \#$\geq$4 & \#$\geq$2 \\
Count & 972 & 136 & 75 & 10 & 33 \\
\midrule
\midrule
T5-base\tiny{(G)}  & 86.2 &  17.5 & 17.5 & 16.7 & 9.0\\
&  92.9&3.0& 5.0& 14.6 & 10.3\\
mBART\tiny{(G)} & 84.8 & 81.5 & 83.1 & 76.6 & 17.2\\
& 91.7 & 80.7 & 84.8 & 80.2 & 17.2\\
mT5\tiny{(G)} & 65.6 & 61.6  & 55.8 & 49.9 & 13.5 \\
&  76.7 & 65.4 & 63.3 &  59.3 & 17.7\\

T5-large\tiny{(G)}  & \textbf{87.9} & 74.8 & 67.4&45.2&19.0\\
&\textbf{94.1} & 73.7 & 68.5& 48.3& 21.5\\
byT5\tiny{(G)}  & 87.3&  \textbf{84.3} & \textbf{86.9} & 52.5 & 29.6\\
& 93.6&85.2& \textbf{86.6}& 55.6 & 33.0\\
 AM Parser\tiny{(G)}  &84.3 & 75.6 & 67.5 &61.9 &46.3 \\
 & 92.1& 78.9&72.0 &66.5& 56.0\\
\midrule
\midrule
AMS Parser\textsubscript{scpl+d}\tiny{(G)}  &86.0 & 83.8 & 81.9 & \textbf{75.2} & \textbf{48.2}\\
& 92.9 & \textbf{86.5} & 82.2 & \textbf{85.2} & \textbf{58.7}\\
\midrule
\midrule
byT5\tiny{(G+S)} & 89.1 & 89.9 & 88.8 & 83.6 & 48.0\\
& 95.0 & 89.2 & 90.6 & 87.4 & 47.8 \\
mT5\tiny{(G+S)} & 89.1 & 89.9 & 88.8 & 85.6 & 61.7\\
& 95.0 & 88.9 & 89.5 & 88.4 & 65.1 \\
mBART\tiny{(G+S)} & 84.8 & 81.5  & 83.1 &75.7 & 17.2 \\

& 91.7 & 80.6 & 84.6 & 80.2 & 17.6\\

\bottomrule
\end{tblr}
\caption{SMATCH score for multi-box DRGs and corresponding scope score (highlighted in gray)}
\label{table:multibox_scope}

\end{table}

At the same time, we observe that the AMS parser maintains a very high accuracy on predicting scope edges even for complex DRGs. We observe that the difference between the AMS Parser and the baseline AM parser is small on single-box DRGs, but much larger on multi-box DRGs, showing that treating scope prediction separately pays off. 

\subsection{What makes long texts so hard?}
\citet{anil2022exploring} found that simple fine-tuning of transformer models does not achieve length generalization, nor does scaling up the models. We conducted a detailed error analysis and identified two factors that might contribute to the limitations of the models in length generalization.

\paragraph{Structural Complexity} As shown in Table~\ref{table:multibox_scope}, all models show a decreasing trend as the number of boxes increases. We find that a higher number of boxes generally results in longer sequences, especially in the TestLong split - we assume the box complexity brought by longer sequences could be a possible reason for length generalization limitation.

Furthermore, \texttt{byT5} tends to generate shorter sequences, averaging 70 roles and relations in its predictions, in contrast to other models which average approximately 100. This discrepancy underscores \texttt{byT5}'s limitation in handling long texts.

\paragraph{Sense Generalization} Furthermore, longer sentences can introduce new word senses, which have to be predicted as node labels. 25\% senses in the TestLong split are absent in the train split. All models show accuracies lower than 0.33 in predicting unseen senses with the AMS Parser performing the best at this rate. 

%% file: appendix.tex
\newpage
\appendix
\section{Statistics of Parallel Meaning Bank Release 5.1.0}
\label{data}
In our experiment, we excluded all ill-formed DRGs from the gold split of the PMB5.1.0 dataset. Detailed statistics of the modified gold data as well as the silver and bronze splits are presented below.
\begin{table}[h!]
\footnotesize
\centering
\begin{tabular}{cccc|cc}
\toprule
\multicolumn{4}{c|}{\textbf{Gold}} & \textbf{Silver}  &\textbf{Bronze} \\ \midrule
                  \textbf{Train} & \textbf{Dev} & \textbf{Test} & \textbf{TestLong} &  &  \\ \midrule
 9560 & 1195 & 1193 & 40& 146,718 & 141,435 \\ 
\bottomrule
\end{tabular}
\caption{Number of sentences across different splits}
\label{tab:my_label}
\end{table}

\section{Challenges Brought by Scope}
\label{scope_challenge}
In this section, we show that AM-Algebra struggles with even one-box DRG when scope is taken as an argument of the root box, with sentence \textit{The little cat wanted to sleep} as an example. 

\begin{figure}[!th]
        \centering
   \includegraphics[scale=0.18]{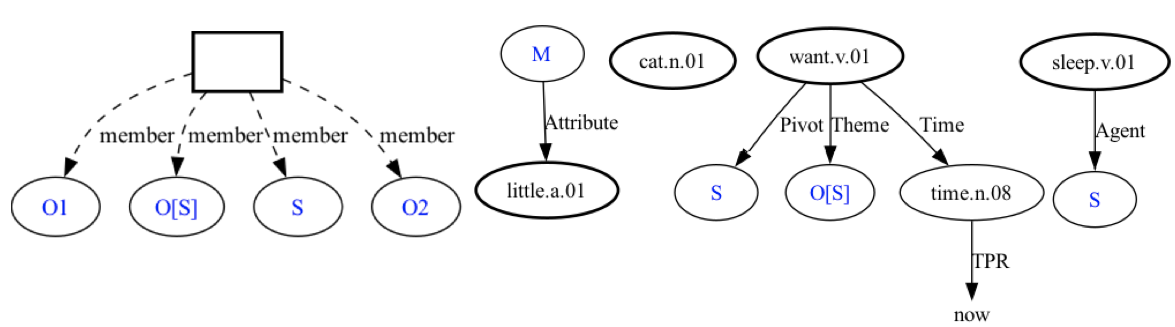}
        \caption{graphs for DRGs with scope as argument}
        \label{drg-challenge}
\end{figure}

The graphs corresponding to the sentences are illustrated in Fig.~\ref{drg-challenge}. As depicted in Fig.~\ref{scopechallenge}, these graphs can be merged to form a scopeless lexical graph. However, integrating this lexical graph with a box requiring four arguments proves problematic for constructing the AM-tree. This is due to AM-Algebra's restriction against multiple \textsc{app}s (applications) between two sub-graphs, a constraint that mirrors linguistic principles in English, where different parts of one constituent cannot play unique roles relative to another constituent.

\begin{figure}[!ht]
    \centering
      \begin{dependency}[theme = simple]
   \begin{deptext}[column sep=.01cm, font=\tiny]
    \textsc{start} \& The \& little \& cat \& wanted \& to \& sleep. \\
    \colorbox{green!20}{\textsc{g}\textsubscript{\textsc{start}}} \& \colorbox{gray!20}{$\bot$}\& \colorbox{red!20}{\textsc{g}\textsubscript{little}} \&   \colorbox{yellow!20}{\textsc{g}\textsubscript{cat}} \&  \colorbox{blue!20}{\textsc{g}\textsubscript{wanted}} \& \colorbox{gray!20}{$\bot$} \& \colorbox{orange!20}{\textsc{g}\textsubscript{sleep}} \\
   \end{deptext}
   \depedge{1}{5}{\textcolor{red}{x}\textsc{mod}\textsubscript{o1}}
   \depedge{5}{4}{\colorbox{blue!20}{\textsc{app}\textsubscript{s}}}
   \depedge{5}{7}{\colorbox{blue!20}{\textsc{app}\textsubscript{o}}}
   \depedge{4}{3}{\colorbox{yellow!20}{\textsc{mod}\textsubscript{m2}}}
\end{dependency}

    \caption{Failed combinations of graphs}
    \label{scopechallenge}
\end{figure}

\section{Heuristics on Edge Directions}
\label{direction}
The heuristics on edge directions can be found in Table~\ref{tab:operation_edge_labels}.
\begin{table}[!th]
\small
    \centering
    \begin{tabular}{p{1cm}p{5.5cm}}
    \toprule
        Operation & Edge Labels \\
        \midrule
        APP & Agent, Bearer, Participant, Creator, Proposition, Stimulus, Beneficiary, Co-Agent, Co-Patient, Co-Theme, Experiencer, Patient, Pivot, Product, Recipient, Theme, Owner, OF, User, Role, NEQ, APX, EQU, TPR\\
        Mod & Consumer, Topic, Result, member, Sub, Source, Destination, Goal, Product, ALTERNATION, ATTRIBUTION, CONDITION, CONSEQUENCE, CONTINUATION, CONTRAST, EXPLANATION, NECESSITY, NEGATION, POSSIBILITY, PRECONDITION, RESULT, SOURCE\\
        \bottomrule
    \end{tabular}
    \caption{Mapping of operation types to edge labels in the DRG-to-graph conversion process.}
    \label{tab:operation_edge_labels}
\end{table}

\section{Coreference Resolution}
\label{coref}
PMB5.1.0 explicitly marks coreference: two nodes that refer to the same entity are connected with an \texttt{ANA} edge.

In our approach, we leverage the AM Parser's supertagger for coreference resolution. In PMB, node labels are annotated with lexical categories like \texttt{n} (noun), \texttt{a} (adjective), \texttt{r} (adverb), and \texttt{v} (verb), such as \texttt{female.n.02} in Fig.~\ref{fig:coref}. To allow coreference resolution via supertagging, we introduce a new category, denoted as \texttt{p} (pronoun). During preprocessing, this category is assigned to nodes involved in coreference, identified by the \texttt{ANA} edge linking them. For example, Fig.~\ref{fig:coref} shows the resulting penman notation after preprocessing and postprocessing steps. The two nodes, \texttt{s0} and \texttt{s3} (bot labeled \texttt{female.n.01}) are relabeled as \texttt{female.p.01}. This encodes the fact that the two entities corefer is now encoded in the node labels, allowing us to remove the \texttt{ANA} edge. While this is not always a lossless transformation when there are multiple instances of coreference in the graph, we find it to work well in practice (see Section~\ref{result}). And crucially, this removes a reentrancy from the DRG, making it more likely to be decomposable by the AM algebra. At training time, the AM Parser's supertagger can then learn to distinguish regular nouns (i.e., \texttt{n}) and coreferent nouns (i.e., \texttt{p}).

\begin{figure}
    \small
        \textbf{Preprocessing}:\\
        \texttt{(b0 / box
	:member (s1 / unscrew.v.01
		:Agent (s0 / female.\st{n}\underline{\textcolor{blue}{\textit{p}}}.02
			:Name (c0 / "Mary"))
		:Time (s2 / time.n.08
			:TPR (c1 / "now"))
		:Patient (s4 / lipstick.n.01
			:User (s3 / female.\st{n}\underline{\textcolor{blue}{\textbf{p}}}}.02  \st{:ANA s0}))))\\
   \textbf{Postprocessing}:\\
   \texttt{(b0 / box
	:member (s1 / unscrew.v.01
		:Agent (s0 / female.\underline{\textcolor{blue}{\textbf{n}}}.02
			:Name (c0 / "Mary")))
		:Time (s2 / time.n.08
			:TPR (c1 / "now"))
		:Patient (s4 / lipstick.n.01
			:User (s3 / female.\underline{\textcolor{blue}{\textbf{n}}}.02 \underline{\textcolor{blue}{\textbf{:ANA s0}}}
   ))))}
    \caption{An example of coreference after preprocessing and postprocessing for the sentence \textit{She$_i$ unscrewed her$_i$ lipstick.}}
    \label{fig:coref}
\end{figure}

At evaluation time, we reconstruct coreference information in a postprocessing step. This step begins with identifying nodes marked as \texttt{p} in predicted DRGs. However, if a DRG contains only one such \texttt{p}-tagged node, we do not treat it as coreferent, since coreference involves multiple entities. In most cases, the parser flags either one or two nodes as potential coreference candidates within a single DRG. When two nodes are both tagged as \texttt{p}, we compare their node concepts to see if they are identical. In our example (Fig.~\ref{fig:coref}), since both nodes are labeled \texttt{female.p.02}, indicating a match, we create an \texttt{ANA} edge linking them. This edge is directed from node with a larger number on the node label (like \texttt{s3}) to the one with a smaller node label (like \texttt{s1}). The final step is to change the nodes' categories from \texttt{p} back to \texttt{n}. 

\section{Implementation details of the scope dependency parser}
\label{appendix:scopeDependencyParser}
The original implementation of \citet{dozat-manning-2018-simpler} uses POS tags, lemma-, and character-level word embeddings, processed through a BiLSTM and a Forward Network (FNN), to predict if there is an edge between two tokens as well as the corresponding edge label. Then a biaffian classifier is used to predict the existence of an edge and the edge label. 

In our experiment, we fine-tune \texttt{roberta-large} \citep{DBLP:journals/corr/abs-1907-11692} and take POS tags and characters as feature embeddings. All the linguistic information is provided by spaCy\footnote{We use version 3.7.2} \cite{honnibal2020spacy}. We keep all other hyperparameters the same as the best model reported in their paper.

\section{Scope Annotation of a Complex Example}
As discussed in Section~\ref{method}, when a single token aligns with a lexical graph that contains multiple nodes or boxes, it creates a complex scenario where different nodes within the same lexical graph are linked to distinct boxes and complicates the establishment of straightforward one-to-one dependency relations between tokens. Our annotation method is straightforward: as long as an aligned lexical graph contains multiple nodes or boxes, we make the scope assignment of each node explicit in a top-down order.  

We illustrate our method with two other possibilities when we build the dependency edges between lexical graphs aligned with tokens.

(1) the two lexical graphs aligned with the token have multiple nodes and boxes respectively, and each node is assigned a different scope box. An example can be found in the scope assignment between the lexical graph aligned with \textit{born} ($G\textsubscript{born}$) and the lexical graph aligned with \textit{all} ($G\textsubscript{all}$). We can see that the bottom node of $G\textsubscript{born}$ receives the scope from the top box of $G\textsubscript{all}$, while the top node of $G\textsubscript{born}$ receives the scope from the bottom box of $G\textsubscript{all}$. In this case, the dependency edge between \textit{born} and \textit{all} is \texttt{scope\_b3\_b2}.

(2) the two lexical graphs aligned with the token have multiple nodes and boxes respectively, and each node is assigned the same scope box. This case can be found in the scope assignment between the lexical graph aligned with \textit{children} ($G\textsubscript{children}$)and that with \textit{all}. Although both nodes of $G\textsubscript{children}$ receives the same scope, we still explicitly annotating the scope for each node as shown in \texttt{scope\_b2\_b2}.

\label{scope_annotation}
\begin{figure}[!th]
\centering
    \begin{subfigure}{0.48\textwidth}
        \centering
\includegraphics[scale=0.28]{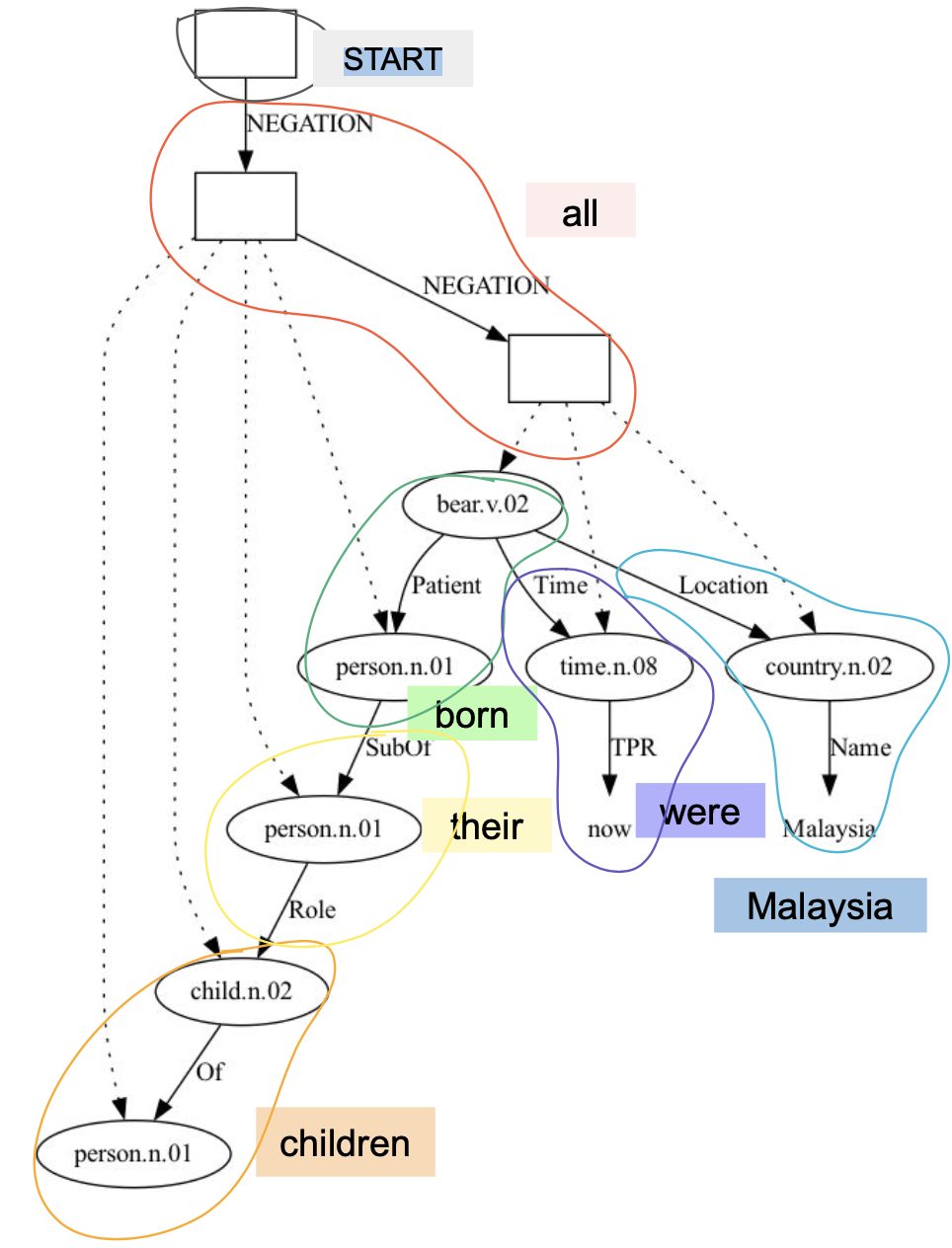}
\captionsetup{font=scriptsize}
\caption{Token Alignments: lexical graphs are color-coded to indicate alignment with distinct tokens, denoted beneath each respective circle}
        \label{fig:complex-drg_align}
    \end{subfigure}
        \begin{subfigure}{0.48\textwidth}
    \centering
        \begin{dependency}[theme = simple]
   \begin{deptext}[column sep=.01cm, font=\tiny]
   \textsc{start} \& All \& of \& their \& children \& were \& born \& in \& Malaysia.\\
    \colorbox{gray!20}{\textsc{g}\textsubscript{\textsc{start}}} \&  \colorbox{red!20}{\textsc{g}\textsubscript{all}} \& of \& 
    \colorbox{yellow!20}{\textsc{g}\textsubscript{their}} 
 \& \colorbox{orange!20}{\textsc{g}\textsubscript{children}} \&   \colorbox{purple!20}{\textsc{g}\textsubscript{were}} \&  \colorbox{green!20}{\textsc{g}\textsubscript{born}}\&
 in \& \colorbox{blue!20}{\textsc{g}\textsubscript{Maylaysia}}\\
   \end{deptext}
   \depedge{1}{3}{no\_scope}
   \depedge{1}{8}{no\_scope}
   \depedge{2}{4}{scope\_b2}
   \depedge{2}{5}{scope\_b2\_b2}
   \depedge{2}{6}{scope\_b3}
   \depedge{2}{7}{scope\_b3\_b2}
   \depedge{2}{9}{scope\_b3}
\end{dependency}
\captionsetup{font=scriptsize}
\caption{The converted dependency graph based on the scope information represented in dashed lines}
\label{fig:complex-tree-scope}
    \end{subfigure}
    \caption{Complex DRG and dependency graph for \textit{All of their children were born in Malaysia.}} 
    \label{fig:complex_annoation}
\end{figure} 

\section{Evaluation Format}
\label{evaluation_format}
In evaluation, we use a more compact format following \citet{wang-etal-2023-discourse}. This strict format integrates synset nodes' information into a single entity and eliminates variables representing constants, thereby avoiding inflated scores. An example is shown in Fig.~\ref{fig:compare-formats}. 

\begin{figure}[!th]
    \begin{subfigure}{.22\textwidth}
        \footnotesize
        \begin{verbatim}
(b0 / "box"
:member (s0 / "synset"
    :lemma "person"
    :pos "n"
    :sense "01"
    :Name (c0 / "?"))
:member (s1 / "synset"
    :lemma "time"
    :pos "n"
    :sense "08"
    :TPR (c1 / "now"))
:member (s2 / "synset"
    :lemma "male"
    :pos "n"
    :sense "02"
    :Name (c2/"William W"))
:member (s3 / "synset"
    :lemma "defeat"
    :pos "v"
    :sense "01"
    :Co-Agent s0
    :Time s1
    :Agent s2))
        \end{verbatim}
        \caption{Lenient Format used in \cite{poelman-etal-2022-transparent}}
        \label{fig:lenient-format}
    \end{subfigure}%
    \hfill
    \begin{subfigure}{.22\textwidth}
        \footnotesize
        \begin{verbatim}
(b0 / box
:member (s0/person.n.01
  :Name "?")
:member (s1 /time.n.08
  :TPR "now")
:member (s2 / male.n.02
  :Name "William W")
:member (s3 /defeat.v.01
  :Co-Agent s0
  :Time s1
  :Agent s2))
        \end{verbatim}
        \caption{Strict Format}
        \label{fig:strict-format}
    \end{subfigure}
    \caption{Comparison of DRG Representation in Lenient and Strict Formats for the sentence \textit{Who did William Wallace defeat?}}
    \label{fig:compare-formats}
\end{figure}

\section{Hyperparameters in AM Parser}
\label{amppp}
The hyperparameters used in the experiments that show the best performance on the scopeless SBN training data are summarized in Table~\ref{table:hyperparameters}.
\begin{table}[ht!]
\small \centering
\begin{tabular}{ll}
\hline
\textbf{Hyperparameter} & \textbf{Value}\\
\hline
Activation function & tanh\\
Optimizer & Adam\\
Learning rate & 0.001\\
Epochs & 100\\
Early Stopping & 20\\
Dim of lemma embeddings & 64\\
Dim of POS embeddings & 32\\
Dim of NE embeddings & 16\\
Minimum lemma frequency & 7\\
Hidden layers in all MLPs & 1\\
Hidden units in LSTM (per direction) & 256\\
Hidden units in edge existence MLP & 256\\
Hidden units in edge label MLP & 256\\
Hidden units in supertagger MLP & 1024\\
Hidden units in lexical label tagger MLP & 1024\\
Layer dropout in LSTMs & 0.35\\
Recurrent dropout in LSTMs & 0.4\\
Input dropout & 0.35\\
Dropout in edge existence MLP & 0.0\\
Dropout in edge label MLP & 0.0\\
Dropout in supertagger MLP & 0.4\\
Dropout in lexical label tagger MLP & 0.4\\
\hline
\end{tabular}
\caption{Common hyperparameters used in all experiments in AM Parser.}
\label{table:hyperparameters}
\end{table}

\section{More examples of Complex DRGs}
\label{simples}
\begin{figure}[th!]
\centering
\includegraphics[width=0.5\textwidth]{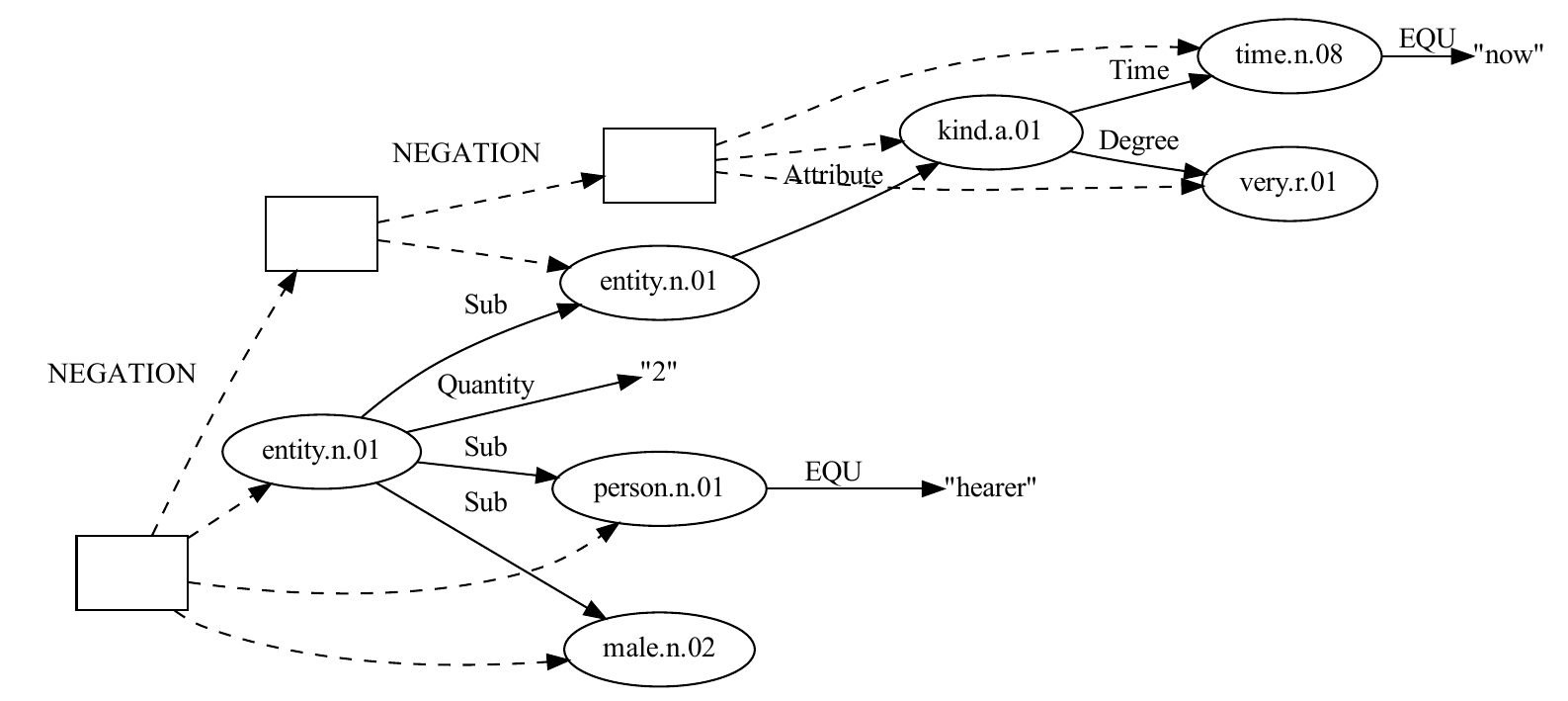}
\hspace{0.001\textwidth}\includegraphics[width=0.5\textwidth]{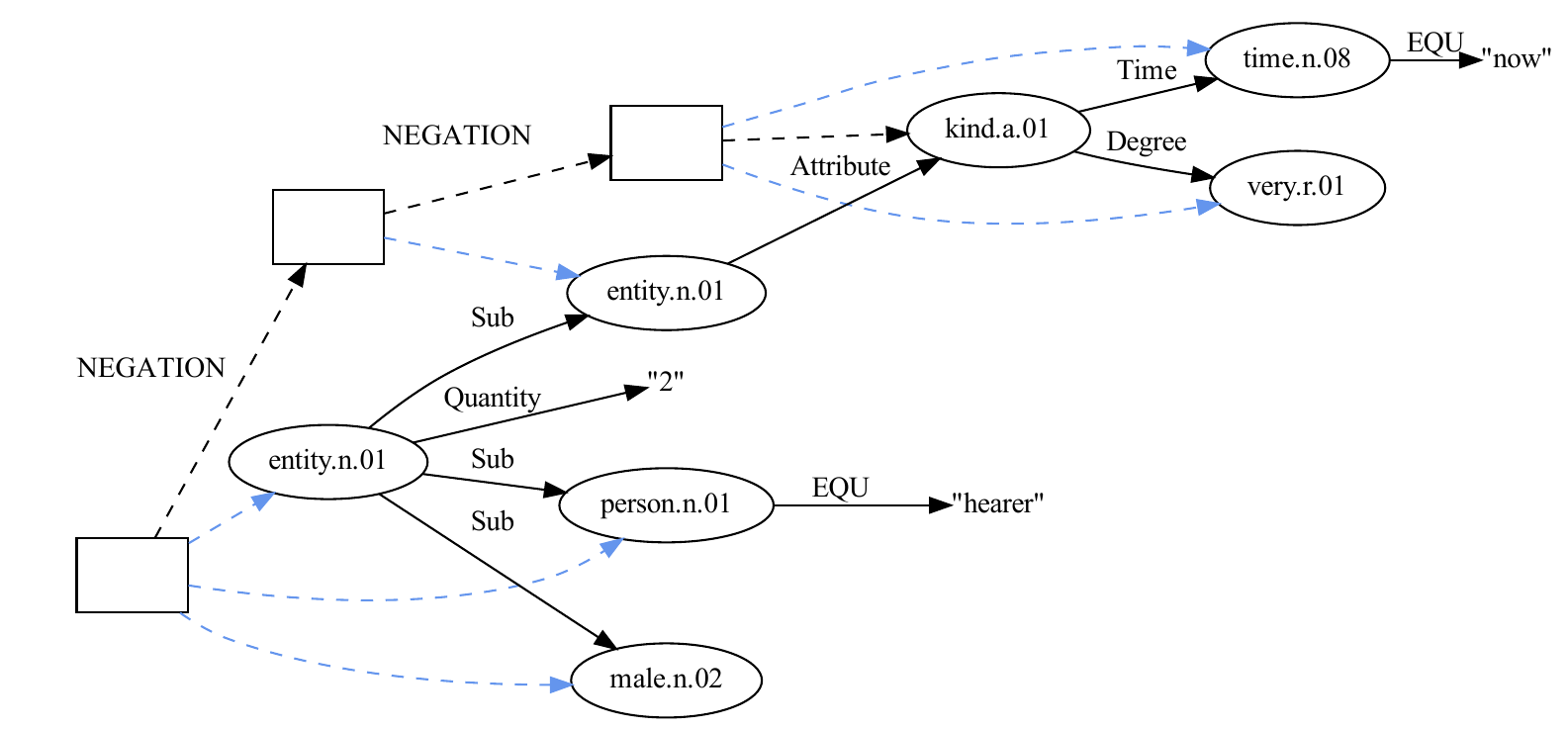}
\hspace{0.001\textwidth}\includegraphics[width=0.5\textwidth]{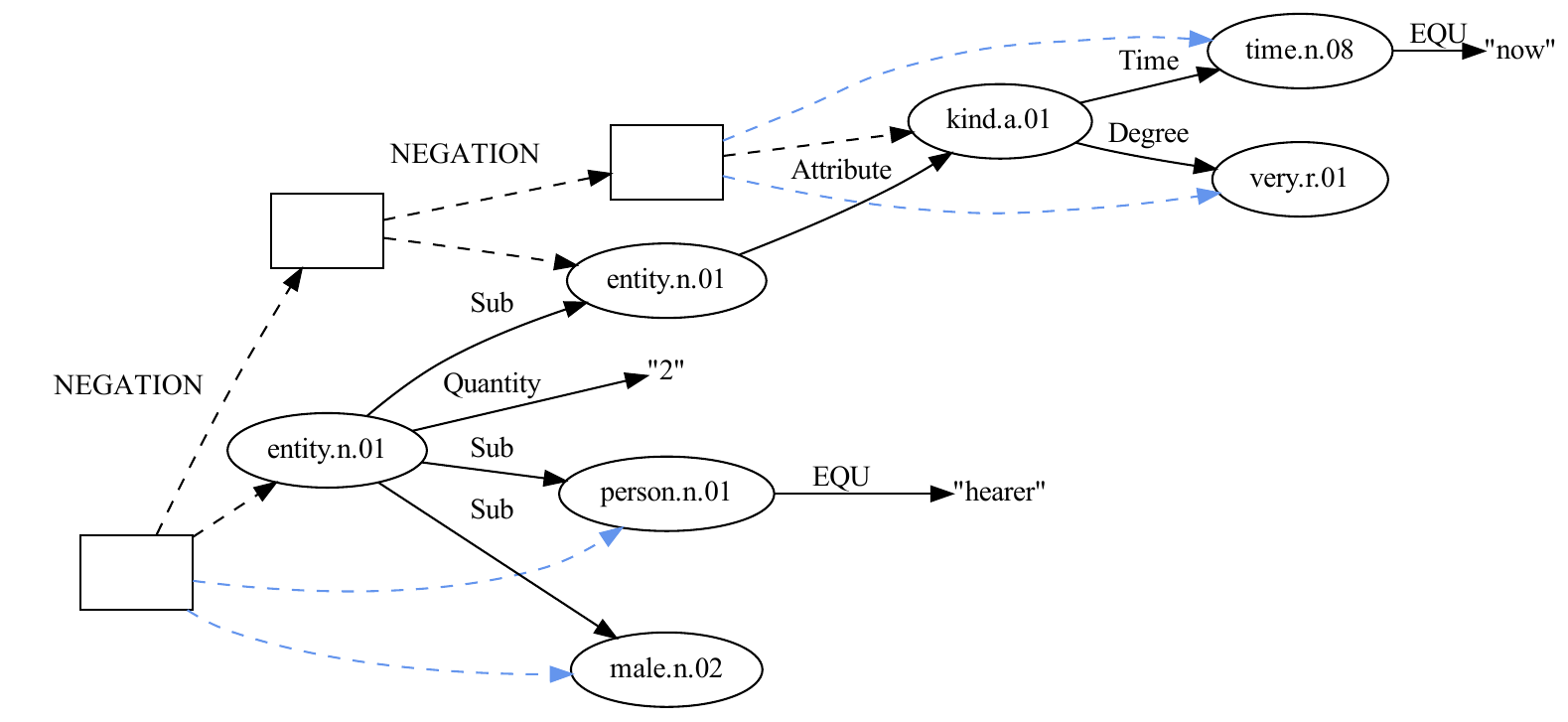}
\caption{Examples of complete DRG (top), scopeless DRG(middle), and simplified DRG (bottom) for the sentence \textit{You and he both are very kind.}}
\label{fig:example2}
\end{figure}

In this section, we show examples of more complex scope assignments in Fig~\ref{fig:example2}.
